\documentclass{article} 
\usepackage{iclr2021_conference,times}

\iclrfinalcopy

\usepackage{amsmath,amsfonts,bm}

\def\eqref#1{equation~\ref{#1}}

\def\1{\bm{1}}

\DeclareMathAlphabet{\mathsfit}{\encodingdefault}{\sfdefault}{m}{sl}
\SetMathAlphabet{\mathsfit}{bold}{\encodingdefault}{\sfdefault}{bx}{n}

\usepackage{hyperref}
\usepackage{url}

\usepackage[utf8]{inputenc} 
\usepackage[T1]{fontenc}    
\usepackage{hyperref}       
\usepackage{url}            
\usepackage{booktabs}       
\usepackage{amsfonts}       
\usepackage{nicefrac}       
\usepackage{microtype}      

\usepackage{hyperref}

\usepackage{amsmath}
\usepackage{amssymb}
\usepackage{amsmath,amsthm,bm,mathrsfs}

\usepackage{tikz}
\usetikzlibrary{positioning}
\usetikzlibrary{decorations.text}
\usetikzlibrary{arrows}
\usetikzlibrary{shapes}
\usetikzlibrary{arrows.meta}

\newtheorem{example}{Example}

\newcommand{\inp}{\mathrm{I}}
\newcommand{\rul}{\mathrm{C}}   
\newcommand{\agg}{\mathrm{A}}

\newcommand{\lab}{\bm{y}}

\usepackage{xcolor}

\tikzstyle{atom}  =  [circle, scale=1.7, shading=ball]
\newcommand\atoms{0}

\tikzset{anchor/.append code=\let\tikz@auto@anchor\relax}
\tikzset{my label/.style args={#1:#2}{
  append after command={
    (\tikzlastnode.center) node [#1] {#2}
    }
  }
}
\newcommand{\wog}{\color{blue} $w_{o_1}$}	
\newcommand{\wogg}{\color{orange} $w_{o_2}$}
\newcommand{\whg}{\color{red} $w_{h_2}$}
\newcommand{\whgg}{$w_{h_1}$}
\newcommand{\wng}{$w_{n_1}$}
\newcommand{\wngg}{$w_{n_2}$}
\newcommand{\ide}{$1$} 

\newcommand{\fweight}{\color{cyan} $w_{f_1}$} 

\tikzstyle{edgenode}  =  [thin, draw=black, align=center,fill=white,font=\small]
\tikzstyle{edgeweight}  =  [near start, above, sloped,outer sep=3pt, inner sep=1pt ,fill=white,font=\large]
\tikzstyle{gredge}  =  [sloped,outer sep=2pt, inner sep=1pt ,fill=white]
\tikzstyle{gredge1}  =  [outer sep=2pt, inner sep=1pt ,fill=white]

\tikzstyle{kappa} = [
  draw, scale = 1.5,  minimum size=1cm, circle, rounded corners,shading=radial,inner color=white,
  minimum height=1cm,
  align=center
  ]
  
\tikzstyle{lambda} = [
  draw, scale = 1.5,  minimum size=1cm, circle, rounded corners,shading=radial,inner color=white,
  minimum height=1cm,
  align=center
  ]

\usepackage{listings}

\definecolor{mygreen}{rgb}{0,0.6,0}
\definecolor{mygray}{rgb}{0.5,0.5,0.5}
\definecolor{kwrds}{rgb}{0.99,0.0,0.0}
\definecolor{backcolour}{rgb}{0.95,0.95,0.92}

\usepackage[noend]{algorithmic}
\usepackage{algorithm}
\usepackage{multicol}

\title{Lossless Compression of Structured \\Convolutional Models via Lifting}

\author{Gustav Sourek, Filip Zelezny, Ondrej Kuzelka\\
    Department of Computer Science\\
    Czech Technical University in Prague\\
    \texttt{\{souregus,zelezny,kuzelon2\}@fel.cvut.cz}
    }

\begin{document}

\maketitle

\begin{abstract}
Lifting is an efficient technique to scale up graphical models generalized to relational domains by exploiting the underlying symmetries. Concurrently, neural models are continuously expanding from grid-like tensor data into structured representations, such as various attributed graphs and relational databases. To address the irregular structure of the data, the models typically extrapolate on the idea of convolution, effectively introducing parameter sharing in their, dynamically unfolded, computation graphs. The computation graphs themselves then reflect the symmetries of the underlying data, similarly to the lifted graphical models. Inspired by lifting, we introduce a simple and efficient technique to detect the symmetries and compress the neural models without loss of any information. We demonstrate through experiments that such compression can lead to significant speedups of structured convolutional models, such as various Graph Neural Networks, across various tasks, such as molecule classification and knowledge-base completion.

\end{abstract}

\section{Introduction}

Lifted, often referred to as~\textit{templated}, models use highly expressive representation languages, typically based in weighted predicate logic, to capture symmetries in relational learning problems~\citep{koller2007introduction}. This includes learning from data such as chemical, biological, social, or traffic networks, and various knowledge graphs, relational databases and ontologies. The idea has been studied extensively in probabilistic settings under the notion of lifted graphical models~\citep{LIFTED}, with instances such as Markov Logic Networks (MLNs)~\citep{MLN} or Bayesian Logic Programs (BLPs)~\citep{BLP}.

In a wider view, \textit{convolutions} can be seen as instances of the templating idea in neural models, where the same parameterized {pattern} is being carried around to exploit the underlying symmetries, i.e. some forms of shared correlations in the data. In this analogy, the popular Convolutional Neural Networks (CNN)~\citep{CNNS} themselves can be seen as a simple form of a templated model, where the template corresponds to the convolutional filters, unfolded over regular spatial grids of pixels. But the symmetries are further even more noticeable in structured, relational domains with discrete element types. With convolutional templates for regular trees, the analogy covers Recursive Neural Networks~\citep{socher2013recursive}, popular in natural language processing. Extending to arbitrary graphs, the same notion covers works such as Graph Convolutional Networks~\citep{kipf2016semi} and their variants~\citep{wu2019comprehensive}, as well as various Knowledge-Base Embedding methods~\citep{wang2017knowledge}. Extending even further to relational structures, there are works integrating parameterized relational logic templates with neural networks~\citep{sourek2018lifted,rocktaschel2017end,marra2019neural,manhaeve2018deepproblog}.

The common underlying principle of templated models is a joint parameterization of the symmetries, allowing for better generalization. However, standard lifted models, such as MLNs, provide another key advantage that, under certain conditions, the model computations can be efficiently carried out without complete template unfolding, often leading to even exponential speedups~\citep{LIFTED}. This is known as ``lifted inference''~\citep{kersting2012lifted} and is utilized heavily in lifted graphical models as well as database query engines~\citep{suciu2011probabilistic}. However, to our best knowledge, this idea has been so far unexploited in the neural (convolutional) models. The main contribution of this paper is thus a ``lifting'' technique to compress symmetries in convolutional models applied to structured data, which we refer to generically as ``structured convolutional models''.



\subsection{Related Work}

The idea for the compression is inspired by lifted inference~\citep{kersting2012lifted} used in templated graphical models. The core principle is that all equivalent sub-computations can be effectively carried out in a single instance and broadcasted into successive operations together with their respective multiplicities, potentially leading to significant speedups. While the corresponding ``liftable'' template formulae (or database queries) generating the isomorphisms are typically assumed to be given~\citep{LIFTED}, we explore the symmetries from the unfolded ground structures, similarly to the approximate methods based on graph bisimulation~\citep{sen2012bisimulation}. All the lifting techniques are then based in some form of first-order variable elimination (summation), and are inherently designed to explore \textit{structural} symmetries in graphical models. In contrast, we aim to additionally explore \textit{functional} symmetries, motivated by the fact that even structurally different neural computation graphs may effectively perform identical function.

The learning in neural networks is also principally different from the model counting-based computations in lifted graphical models in that it requires many consecutive evaluations of the models as part of the encompassing iterative training routine. Consequently, even though we assume to unfold a complete computation graph before it is compressed with the proposed technique, the resulting speedup due to the subsequent training is still substantial. From the deep learning perspective, there have been various model compression techniques proposed to speedup the training, such as pruning, decreasing precision, and low-rank factorization~\citep{cheng2017survey}. However, to our best knowledge, the existing techniques are lossy in nature, with a recent exception of compressing ReLU networks based on identifying neurons with linear behavior~\citep{serra2020lossless}. None of these works exploit the model computation symmetries.
The most relevant line of work here are Lifted Relational Neural Networks (LRNNs)~\citep{sourek2018lifted} which however, despite the name, provide only templating capabilities without lifted inference, i.e. with complete, uncompressed ground computation graphs.

\section{Background}
\label{sec:background}

The compression technique described in this paper is applicable to a number of structured convolutional models, ranging from simple recursive~\citep{socher2013recursive} to fully relational neural models~\citep{sourek2018lifted}. The common characteristic of the targeted learners is the utilization of convolution (templating), where the same parameterized pattern is carried over different subparts of the data (representation) with the same local structure, effectively introducing repetitive sub-computations in the resulting computation graphs, which we exploit in this work.


\subsection{Graph Neural Networks}\label{sec:gnn}

Graph neural networks (GNNs) are currently the most prominent representatives of structured convolutional models, which is why we choose them for brevity of demonstration of the proposed compression technique. GNNs can be seen as an extension of the common CNN principles to completely irregular graph structures. Given a particularly structured input sample graph $\mathrm{S}_j$, they dynamically unfold a multi-layered computation graph $\mathcal{G}_j$, where the structure of each layer $i$ follows the structure of the {whole} input graph $\mathrm{S}_j$. For computation of the next layer $i+1$ values, each node $v$ from the input graph $\mathrm{S}_j$ calculates its own value $h(v)$ by {aggregating} $A$ (``pooling'') the values of the adjacent nodes $u : {edge}(u,v)$, transformed by some parametric function $C_{W_1}$ (``convolution''), which is being reused with the same parameterization $W_1$ within each layer $i$ as:

\begin{equation}
\Tilde{h}(v)^{(i)} = A^{(i)}(\{C^{(i)}_{W_1^i}(h(u)^{(i-1)}) | u : {edge}(u,v) \})
\end{equation}
The $\Tilde{h}^{(i)}(v)$ can be further combined through another $C_{W_2}$ with the central node's representation from the previous layer to obtain the final updated value $h^{(i)}(v)$ for layer $i$ as:
\begin{equation}
h(v)^{(i)} = C_{W_2^i}^{(i)} (h(v)^{(i-1)}, \Tilde{h}(v)^{(i)})
\end{equation}

This general principle covers a wide variety of GNN models, such the popular GCNs~\citep{kipf2016semi}, graph-SAGE~\citep{hamilton2017inductive}, GIN~\citep{xu2018powerful}, and others~\citep{xu2018representation,gilmer2017neural}, which then reduces to the respective choices of particular aggregations $A$ and transformations $C_{W}$.
An example computation graph of a generic GNN unfolded over an example molecule of methane is shown in Fig.~\ref{fig:GNNs}.



\subsection{Computation Graphs}
\label{sec:computation_graph}
For the sake of this paper, let us now define the common notion of a computation graph more formally.
A computation graph is a tuple $\mathcal{G} = (\mathcal{N},\mathcal{E},\mathcal{F})$, where $\mathcal{N} = (1,2,\dots,n)$ is a list of {\em nodes} and $\mathcal{E} \subseteq \mathcal{N}^2 \times \mathbb{N}$ is a list of directed labeled \textit{edges}. Each labeled edge is a triple of integers $(n_1,n_2,l)$ where $n_1$ and $n_2$ are nodes of the computation graph and $l$ is the {\em label}. The labels are used to assign weights to the edges in the computation graph. Note this allows to define the weight sharing scheme as part of the graph (cf Example \ref{ex:simple} below). Finally, $\mathcal{F} = \{f_1,f_2,\dots,f_n \}$ is the list of {\em activation functions}, one for each node from $\mathcal{N}$.
As usual, the graph is assumed to be acyclic. 
Children of a node $N$ are naturally defined as all those nodes $M$ such that $(M,N,L) \in \mathcal{E}$, and analogically for parents. 
Note that since $\mathcal{E}$ is a list, edges contained in it are ordered, and the same edge may appear multiple times (which will be useful later).
Children of each node are also ordered -- given two children $C$ and $C'$ of a node $N$, $C$ precedes $C'$ iff $(C,N,L)$ precedes $(C',N,L')$ in the list of edges $\mathcal{E}$. We denote the lists of children and parents of a given node $N$ by $\textit{Children}(N)$ and $\textit{Parents}(N)$, respectively. Computation graphs are then evaluated bottom up from the leaves of the graph (nodes with no children) to the roots of the graph (nodes with no parents).
Given a list of weights $\mathcal{W}$, we can now define the {\em value} of a node $N \in \mathcal{N}$ recursively as:
$$\textit{value}(N; \mathcal{W}) = f_N \Big( \mathcal{W}_{L_1} \cdot \textit{value}(M_1; \mathcal{W}), \dots, \mathcal{W}_{L_m} \cdot \textit{value}(M_m; \mathcal{W}) \Big),$$
where $(M_1,\dots, M_m) \equiv \textit{Children}(N)$ is the (ordered) list of children of the node $N$, and $L_1, \dots, L_m$ are the labels of the respective edges $(M_1,N,L_1)$, $\dots$, $(M_m,N,L_m) \in \mathcal{E}$, and $\mathcal{W}_{L_i}$ is the $L_i$-th component of the list $\mathcal{W}$. 
Note that with the structured convolutional models, such as GNNs, we assume dynamic computation graphs where each learning sample $\mathrm{S}_j$ generates a separate $\mathcal{G}_j$. Consequently, we can associate the leaf nodes in each $\mathcal{G}_j$ with constant functions\footnote{in contrast to static computation graphs where these functions are identities requiring the features at input.}, outputting the corresponding node (feature) values from the corresponding structured input sample $\mathrm{S}_j$.

\section{Problem Definition}

The problem of detecting the symmetries in computation graphs can then be formalized as follows.

\newtheorem{definition}{Definition}

\begin{definition}[Problem Definition]\label{def:problem}
Let $\mathcal{G} = (\mathcal{N},\mathcal{E},\mathcal{F})$ be a computation graph. We say that two nodes $N_1, N_2$ are equivalent if, for any $\mathcal{W}$, it holds that $\textit{value}(N_1; \mathcal{W}) = \textit{value}(N_2; \mathcal{W})$. The problem of detecting symmetries in computation graphs asks to partition the nodes of the computation graph into equivalence classes of mutually equivalent nodes.
\end{definition}


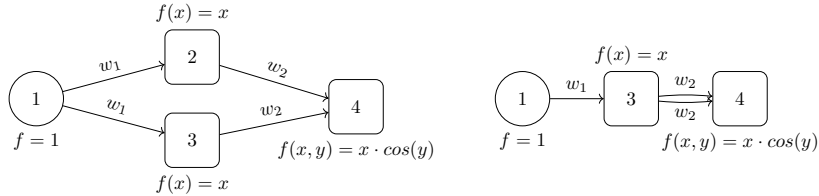
\begin{figure*}[]
\centering
\resizebox{0.8\textwidth}{!}{
	\begin{tikzpicture}
	  \node[scale=1] (rec1) at (0,0){\begin{tikzpicture}
[transform shape,rotate=0, node distance=2.0cm and 2.0cm,
ar/.style={->,>=latex},
mynode/.style={
  draw, scale = 1.0,  minimum size=1cm, rounded corners,left color=white,
  minimum height=1cm,
  align=center
  }
]

\tikzstyle{recnode}  =  [mynode, right color=white!30!white]

\node[recnode, circle, label={below:{$f=1$}}] (n1) {$1$};

\node[recnode, label={above:{$f(x)=x$}}] (n2) [above right = -0.1cm and 2cm of n1] {$2$};
\node[recnode, label={below:{$f(x)=x$}}] (n3) [below right = -0.1cm and 2cm of n1] {$3$};

\node[recnode, label={below:{$f(x,y)=x\cdot cos(y)$}}] (n4) [below right = -0.1cm and 2cm of n2] {$4$};

\draw[->] (n1) -> node[above,sloped] {$w_1$} (n2);
\draw[->] (n1) -> node[above,sloped] {$w_1$} (n3);

\draw[->] (n3) -> node[above,sloped] {$w_2$} (n4);
\draw[->] (n2) -> node[above,sloped] {$w_2$} (n4);

\end{tikzpicture}};
	  \node[scale=1] (rec2) at (8,0){\begin{tikzpicture}
[transform shape,rotate=0, node distance=2.0cm and 2.0cm,
ar/.style={->,>=latex},
mynode/.style={
  draw, scale = 1.0,  minimum size=1cm, rounded corners,left color=white,
  minimum height=1cm,
  align=center
  }
]

\tikzstyle{recnode}  =  [mynode, right color=white!30!white]

\node[recnode, circle, label={below:{$f=1$}}] (n1) {$1$};

\node[recnode, label={above:{$f(x)= x$}}] (n2) [right of = n1] {${3}$};

\node[recnode, label={below:{$f(x,y)=x\cdot cos(y)$}}] (n4) [right of = n2] {$4$};

\draw[->] (n1) -> node[above] {$w_1$} (n2);

\draw[->] (n2) to[out=-5,in=-175] node[below] {$w_2$} (n4);
\draw[->] (n2) to[out=5,in=175] node[above] {$w_2$} (n4);

\end{tikzpicture}};
	\end{tikzpicture}
}
\caption{Depiction of the computation graph (left) compression (right) from Example~\ref{ex:simple}.}
\label{fig:simple}
\end{figure*}


\begin{example}
\label{ex:simple}
Consider the computation graph $\mathcal{G} = (\mathcal{N},\mathcal{E},\mathcal{F})$, depicted in Fig.~\ref{fig:simple}, where
\begin{align*}
    \mathcal{N} &= \{1,2,3,4\},\; 
    \mathcal{E} = ((1,2,1),(1,3,1),(2,4,2),(3,4,2)), \\
    \mathcal{F} &= \{ f_1 = 1, f_2(x) = f_3(x) = x, f_4(x,y) = x\cdot cos(y) \}.
\end{align*}
Let $\mathcal{W} = (w_1,w_2)$ be the weight list. The computation graph then computes the function $(w_1 w_2)\cdot cos{(w_1 w_2)}$. It is not difficult to verify that the nodes $2$ and $3$ are functionally equivalent. This also means, as we discuss in more detail in the next section, that we can ``replace'' $2$ with $3$ without changing the function that the graph computes. The resulting reduced graph then has the form
\begin{align*}
    \mathcal{N} &= \{1,3,4\}, \;
    \mathcal{E} = \{(1,3,1),(3,4,2),(3,4,2)\},  \\
    \mathcal{F} &= \{ f_1 = 1, f_3(x) = x, f_4(x,y) = x\cdot cos(y) \}.
\end{align*}
\end{example}

In the example above, the nodes $2$ and $3$ are in fact also isomorphic in the sense that there exists an automorphism (preserving weights and activation functions) of the computation graph that swaps the two nodes. Note that our definition is less strict: all we want the nodes to satisfy is {\em functional} equivalence, meaning that they should evaluate to the same values for any initialization of $\mathcal{W}$.

{We will also use the notion of {\em structural-equivalence} of nodes in computational graphs. Two nodes are structurally equivalent if they have the same outputs for any assignment of weights $\mathcal{W}$ and for any replacement of any of the activation functions in the graph.\footnote{Here, we add that in this definition, obviously, when we replace a function $f$ by function $f'$, we have to replace all occurrences of $f$ in the graph also by $f'$.} That is if two nodes are structurally equivalent then they are also functionally equivalent but not vice versa. Importantly, the two nodes do not need to be automorphic\footnote{Here, when we say ``automorphic nodes'', we mean that there exists an automorphism of the graph swapping the two nodes.} in the graph-theoretical sense while being structurally equivalent, which also makes detecting structural equivalence easier from the computational point of view. In particular, we describe a simple polynomial-time algorithm in Section \ref{sec:exact-algorithm}.}



\section{Two Algorithms for Compressing computation Graphs}
\label{sec:lifting}


In this section we describe two algorithms for compression of computation graphs: a non-exact algorithm for compression based on functional equivalency (cf. Definition \ref{def:problem}) and an exact algorithm for compression based on detection of structurally-equivalent nodes in the computation graph. While the exact algorithm will guarantee that the original and the compressed computation graphs represent the same function, that will not be the case for the non-exact algorithm. Below we first describe the non-exact algorithm and then use it as a basis for the exact algorithm. 

\subsection{A Non-Exact Compression Algorithm}

The main idea behind the non-exact algorithm is almost embarrassingly simple. The algorithm first evaluates the computation graph with $n$ randomly sampled parameter lists $\mathcal{W}_1$, $\dots$, $\mathcal{W}_n$, i.e.~with $n$ random initializations of the (shared) weights, and records the values of all the nodes of the computation graph (i.e.\ $n$ values per node). It then traverses the computation graph from the output nodes in a breadth-first manner, and whenever it processes a node $N$, for which there exists a node $N'$ that has not been processed yet and all $n$ of its recorded values are the same as those of the currently processed node $N$, the algorithm replaces $N$ by $N'$ in the computation graph. In principle, using larger $n$ will decrease the probability of merging nodes that are not functionally equivalent as long as there is a non-zero chance that any two non-equivalent nodes will have different values (this is the same as the ``amplification trick'' normally used in the design of randomized algorithms).


It is easy to see that any functionally equivalent nodes will be mapped by the above described algorithm to the same node in the compressed computation graph. However, it can happen that the algorithm will also merge nodes that are not functionally equivalent but just happened (by chance) to output the same values on all the random parameter initializations that the algorithm used. We acknowledge that this can happen in practice, nevertheless it was not commonly encountered in our experiments (Sec.~\ref{sec:experiments}), unless explicitly emulated. To do that, we decreased the number of significant digits used in each equivalence check between $\textit{value}(N; \mathcal{W}_i)$ and $\textit{value}(N'; \mathcal{W}_i)$. This allows to compress the graphs even further, at the cost of sacrificing fidelity w.r.t. the original model.


There are also cases when we can give (probabilistic) guarantees on the correctness of this algorithm. One such case is when the activation functions in the computation graph are all polynomial. In this case, we can use DeMillo-Lipton-Schwartz-Zippel Lemma~\citep{demillo1977probabilistic} to bound the probability of merging two nodes that are not functionally equivalent. However, since the activation functions in the computation graphs that we are interested in are usually not polynomial, we omit the details here. In particular, obtaining similar probabilistic guarantees with activation functions such as ReLU does not seem doable.\footnote{In particular, the proof of DeMillo-Lipton-Schwartz-Zippel Lemma relies on the fact that any single variable polynomial is zero for only a finite number of points, which is not the case for computation graphs with ReLUs.}

\begin{figure*}[t]
\centering
\resizebox{1.0\textwidth}{!}{
	\begin{tikzpicture}
	  \node[scale=1] (rec1) at (0,0){\begin{tikzpicture}
[transform shape,rotate=0, node distance=2.0cm and 2.0cm,
ar/.style={->,>=latex},
mynode/.style={
  draw, scale = 1.0,  minimum size=1cm, rounded corners,left color=white,
  minimum height=1cm,
  align=center
  }
]

\tikzstyle{neuron}  =  [rectangle, draw, scale = 1.0,  minimum size=1cm]
\tikzstyle{treenode}  =  [mynode, right color=black!30!white]
\tikzstyle{recnode}  =  [mynode, right color=brown!20!white]
\tikzstyle{recnode2}  =  [mynode, right color=brown!40!white]
\tikzstyle{rulenode}  =  [mynode, right color=red!30!white]
\tikzstyle{aggnode}  =  [mynode, right color=blue!30!white]
\tikzstyle{qnode}  =  [mynode, right color=brown!60!white]

\tikzstyle{grbond}  =  [mynode, right color=black!30!white]
\tikzstyle{gratom}  =  [mynode]
\tikzstyle{grgroup} =  [mynode, right color=brown!30!white]
\tikzstyle{grexpl}  =  [mynode, right color=violet!30!white]
\tikzstyle{edgenode}  =  [thin, draw=black, align=center,fill=white,font=\small]

\node[mynode, left color=black!60!white] (g1) {$\inp_{\rm{h}(c_1)}$};
\node[treenode] (g2) [below left = 0.55 and 0.45cm of g1] {$\inp_{\rm{h}(h_1)}$};
\node[treenode] (g3) [below right = 0.55 and 0.45cm of g1] {$\inp_{\rm{h}(h_2)}$};
\node[treenode] (g4) [above left = 0.55 and 0.45cm of g1]{$\inp_{\rm{h}(h_3)}$};
\node[treenode] (g5) [above right = 0.55 and 0.45cm of g1]{$\inp_{\rm{h}(h_4)}$};

\draw[black,-] (g1) -> (g2);
\draw[black,-] (g1) -> (g3);
\draw[black,-] (g1) -> (g4);
\draw[black,-] (g1) -> (g5);

\node[recnode] (h1) [right = 6.5cm of g1] {$\rul_{\rm{h}(c_1)}^{(1)}$};
\node[recnode] (h2) [below left = 0.5 and 0.5cm of h1] {$\rul_{\rm{h}(h_1)}^{(1)}$};
\node[recnode] (h3) [below right = 0.5 and 0.5cm of h1] {$\rul_{\rm{h}(h_2)}^{(1)}$};
\node[recnode] (h4) [above left = 0.5 and 0.5cm of h1] {$\rul_{\rm{h}(h_3)}^{(1)}$};
\node[recnode] (h5) [above right = 0.5 and 0.5cm of h1] {$\rul_{\rm{h}(h_4)}^{(1)}$};

\node[rulenode,scale=0.5] (r1) [below right=2 and 4.8cm of g4] {$\rul^4_1$};
\node[rulenode,scale=0.5] (r6) [above right=2 and 4.8cm of g2] {$\rul^2_1$};

\node[rulenode,scale=0.5] (r2) [below right=-2.2 and 2.5cm of g1] {$\rul^1_4$};
\node[rulenode,scale=0.5] (r3) [below right=-1.25 and 2.5cm of g1] {$\rul^1_2$};
\node[rulenode,scale=0.5] (r4) [below right=-0.25 and 2.5cm of g1] {$\rul^1_3$};
\node[rulenode,scale=0.5] (r5) [below right= 0.7 and 2.5cm of g1] {$\rul^1_5$};

\node[rulenode,scale=0.5] (r7) [above right=0.2 and 3.8cm of g5] {$\rul^3_1$};
\node[rulenode,scale=0.5] (r8) [below right=0.2 and 3.8cm of g3] {$\rul^5_1$};

\node[aggnode,scale=0.5] (rh1) [below left = -0.8cm and 2cm of h1] {$\agg$1};
\node[aggnode,scale=0.5] (rh2) [right= 0.3cm of r6] {$\agg$4};
\node[aggnode,scale=0.5] (rh3) [below left = -0.2cm and 0.5cm of h3] {$\agg$3};
\node[aggnode,scale=0.5] (rh4) [right= 0.3cm of r1] {$\agg$2};
\node[aggnode,scale=0.5] (rh5) [above left = -0.2cm and 0.5cm of h5] {$\agg$5};

\draw[gray,->] (r1) -> (rh4);
\draw[gray,->] (rh4) -> (h2);

\draw[gray,->] (r8) -> (rh3);
\draw[gray,->] (rh3) -> (h3);

\draw[gray,->] (r2) -> (rh1);
\draw[gray,->] (r3) -> (rh1);
\draw[gray,->] (r4) -> (rh1);
\draw[gray,->] (r5) -> (rh1);
\draw[gray,->] (rh1) -> (h1);

\draw[gray,->] (r7) -> (rh5);
\draw[gray,->] (rh5) -> (h5);

\draw[gray,->] (r6) -> (rh2);
\draw[gray,->] (rh2) -> (h4);

\draw[black,dotted] (h1) -> (h2);
\draw[black,dotted] (h1) -> (h3);
\draw[black,dotted] (h1) -> (h5);
\draw[black,dotted] (h1) -> (h4);

\draw[violet,->] (g1) to[out=2,in=-168] node[below, near end] {$W_2$} (h1);
\draw[violet,->] (g2) to[out=-20,in=-170] node[below, near start] {$W_2$} (h2);
\draw[violet,->] (g4) to[out=20,in=170] node[ above, near start] {$W_2$} (h4);

\draw[violet,->] (g3) to[out=-20,in=-140] node[ below, near start] {$W_2$} (h3);
\draw[violet,->] (g5) to[out=20,in=140] node[ above, near start] {$W_2$} (h5);

\draw[orange,->] (g1) -> node[above] {$W_1$} (r1);
\draw[orange,->] (g1) -> node[below] {$W_1$} (r6);
\draw[orange,->] (g1) to[out=21,in=170] node[below, near end] {$W_1$} (r7);
\draw[orange,->] (g1) to[out=-21,in=-170] node[above, near end] {$W_1$} (r8);

\draw[orange,->] (g2) -> node[below, near start] {$W_1$} (r4);
\draw[orange,->] (g3) -> node[above, near end] {$W_1$} (r5);
\draw[orange,->] (g4) -> node[below, near start] {$W_1$} (r3);
\draw[orange,->] (g5) -> node[below, near end] {$W_1$} (r2);

\node[recnode2] (hh1) [right = 5.5cm of h1] {$\rul_{\rm{h}(c_1)}^{(n)}$};
\node[recnode2] (hh2) [below left = 0.5 and 0.5cm of hh1] {$\rul_{\rm{h}(h_1)}^{(n)}$};
\node[recnode2] (hh3) [below right = 0.5 and 0.5cm of hh1] {$\rul_{\rm{h}(h_2)}^{(n)}$};
\node[recnode2] (hh4) [above left = 0.5 and 0.5cm of hh1] {$\rul_{\rm{h}(h_3)}^{(n)}$};
\node[recnode2] (hh5) [above right = 0.5 and 0.5cm of hh1] {$\rul_{\rm{h}(h_4)}^{(n)}$};

\draw[black,loosely dotted] (hh1) -> (hh2);
\draw[black,loosely dotted] (hh1) -> (hh3);
\draw[black,loosely dotted] (hh1) -> (hh4);
\draw[black,loosely dotted] (hh1) -> (hh5);

\draw[cyan,dashed,->] (h1) -> node[color=black, draw=white, fill=white] {\Large$\dots$} (hh1);
\draw[cyan,dashed,->] (h2) to[out=-40,in=-160] node[color=black, draw=white, fill=white] {\Large$\dots$} (hh2);
\draw[cyan,dashed,->] (h3) to[out=-20,in=-150] node[color=black, draw=white, fill=white] {\Large$\dots$} (hh3);
\draw[cyan,dashed,->] (h4) to[out=40,in=160] node[color=black, draw=white, fill=white] {\Large$\dots$} (hh4);
\draw[cyan,dashed,->] (h5) to[out=20,in=150] node[color=black, draw=white, fill=white] {\Large$\dots$} (hh5);

\draw[blue,dashed,->] (h1) -> node[color=black, draw=white, fill=white] {\Large$\dots$} (hh2);
\draw[blue,dashed,->] (h1) -> node[color=black, draw=white, fill=white] {\Large$\dots$} (hh3);
\draw[blue,dashed,->] (h1) -> node[color=black, draw=white, fill=white] {\Large$\dots$} (hh4);
\draw[blue,dashed,->] (h1) -> node[color=black, draw=white, fill=white] {\Large$\dots$} (hh5);

\draw[blue,dashed,->] (h2) -> node[color=black, draw=white, fill=white] {\Large$\dots$} (hh1);
\draw[blue,dashed,->] (h3) -> node[color=black, draw=white, fill=white] {\Large$\dots$} (hh1);
\draw[blue,dashed,->] (h4) -> node[color=black, draw=white, fill=white] {\Large$\dots$} (hh1);
\draw[blue,dashed,->] (h5) -> node[color=black, draw=white, fill=white] {\Large$\dots$} (hh1);

\node[aggnode] (agg) [right = 2cm of hh1] {$\agg^{\lab}$};
\draw[gray,->] (hh1) -> (agg);
\draw[gray,->] (hh2) -> (agg);
\draw[gray,->] (hh3) -> (agg);
\draw[gray,->] (hh4) -> (agg);
\draw[gray,->] (hh5) -> (agg);

\node[qnode] (q) [right = 1.5cm of agg] {$\rul^{\lab}$};
\draw[brown,->] (agg) -> node[below] {$W^{(n)}$} (q);

\end{tikzpicture}};
	  \node[scale=0.4] (met) at (9.7,-1.8){

\begin{tikzpicture}
[transform shape,rotate=0, node distance=2.0cm and 2.0cm,
ar/.style={->,>=latex},
mynode/.style={
  draw, scale = 1.0,  minimum size=1cm, rounded corners,
  minimum height=1cm,
  align=center
  }
]

\node[atom, ball color=red!50!white] at (0,1) (o1) {\Large \textbf{C\textsubscript{{\tiny 1}}}};

\node[atom, circle, ball color=blue!50!white] at (1,0) (h1) {\textbf{H\textsubscript{\tiny 2}}};
\node[atom, circle, ball color=blue!50!white] at (1,2) (h2) {\textbf{H\textsubscript{\tiny 4}}};

\node[atom, circle, ball color=blue!50!white] at (-1,0) (h2) {\textbf{H\textsubscript{\tiny 1}}};
\node[atom, circle, ball color=blue!50!white] at (-1,2) (h2) {\textbf{H\textsubscript{\tiny 3}}};

\node[mynode, dashed] at (4,1) (q) {\Large\textbf{$\lab$}};

\draw[black, -{Latex[length=5mm, width=4mm]}, dashed, shorten <=0.9cm] (o1) -> (q);

\end{tikzpicture}};
	\end{tikzpicture}
}
\caption{A multi-layer GNN model with a global readout unfolded over an example~molecule~of~methane. Colors are used to distinguish the weight sharing, as well as different node types categorized w.r.t. the associated activation functions, denoted as input ($\inp$), convolution ($\rul$), and aggregation ($\agg$) nodes, respectively.}
\label{fig:GNNs}
\end{figure*}
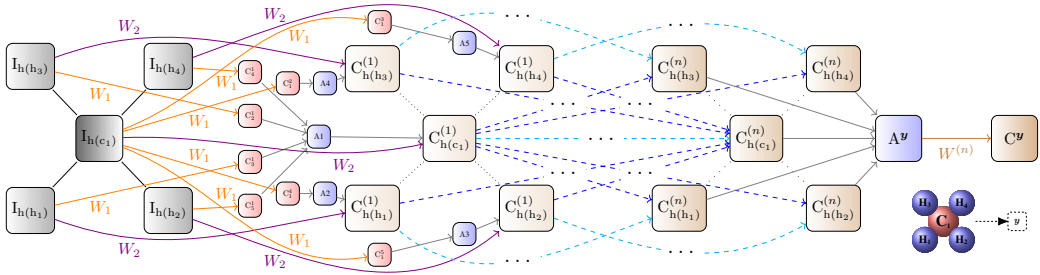

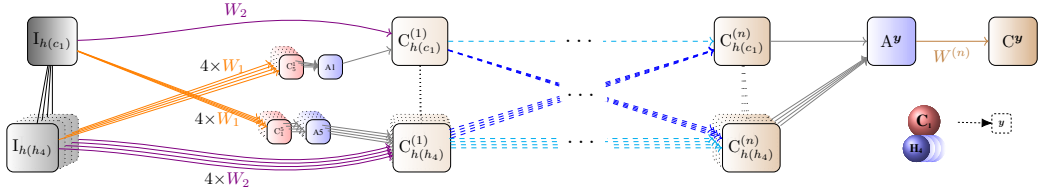
\begin{figure*}[]
\centering
\resizebox{1.0\textwidth}{!}{
	\begin{tikzpicture}
	  \node[scale=1] (rec1) at (0,0){\begin{tikzpicture}
[transform shape,rotate=0, node distance=2.0cm and 2.0cm,
ar/.style={->,>=latex},
mynode/.style={
  draw, scale = 1.0,  minimum size=1cm, rounded corners,left color=white,
  minimum height=1cm,
  align=center
  }
]

\newcommand{\dist}{-0.95cm}	
\newcommand{\dista}{-1.15cm}	
\newcommand{\distb}{-.45cm}	
\newcommand{\distc}{-1.2cm}	

\tikzstyle{neuron}  =  [rectangle, draw, scale = 1.0,  minimum size=1cm]
\tikzstyle{treenode}  =  [mynode, right color=black!30!white]
\tikzstyle{recnode}  =  [mynode, right color=brown!20!white]
\tikzstyle{recnode2}  =  [mynode, right color=brown!40!white]
\tikzstyle{rulenode}  =  [mynode, right color=red!30!white]
\tikzstyle{aggnode}  =  [mynode, right color=blue!30!white]
\tikzstyle{qnode}  =  [mynode, right color=brown!60!white]

\tikzstyle{grbond}  =  [mynode, right color=black!30!white]
\tikzstyle{gratom}  =  [mynode]
\tikzstyle{grgroup} =  [mynode, right color=brown!30!white]
\tikzstyle{grexpl}  =  [mynode, right color=violet!30!white]
\tikzstyle{edgenode}  =  [thin, draw=black, align=center,fill=white,font=\small]

\node[mynode, left color=black!60!white] (g1) {$\inp_{h(c_1)}$};
\node[treenode,dotted] (g2) [below =1cm of g1] {$\inp_{h(n_2)}$};
\node[treenode,dotted] (g3) [below left=\dist and \dist of g2] {$\inp_{h(n_3)}$};
\node[treenode,dotted] (g4) [below left=\dist and \dist of g3]{$\inp_{h(n_4)}$};
\node[treenode] (g5) [below left=\dist and \dist of g4]{$\inp_{h(h_4)}$};

\draw[black,-] (g1) -> (g2);
\draw[black,-] (g1) -> (g3);
\draw[black,-] (g1) -> (g4);
\draw[black,-] (g1) -> (g5);

\node[recnode] (h1) [right = 6.5cm of g1] {$\rul_{h(c_1)}^{(1)}$};
\node[recnode,dotted] (h2) [below=1cm of h1] {$\rul_{h(n_2)}^{(1)}$};
\node[recnode,dotted] (h3) [below right=\dist and \distc of h2] {$\rul_{h(n_3)}^{(1)}$};
\node[recnode,dotted] (h4) [below right=\dist and \distc of h3] {$\rul_{h(n_4)}^{(1)}$};
\node[recnode] (h5) [below right=\dist and \distc of h4] {$\rul_{h(h_4)}^{(1)}$};

\node[rulenode,dotted,scale=0.5] (r6) [below right=-1.2 and 4.cm of g4] {$\rul^2_1$};
\node[rulenode,dotted,scale=0.5] (r7) [below right=\distb and \distb of r6] {$\rul^3_1$};
\node[rulenode,dotted,scale=0.5] (r1) [below right=\distb and \distb of r7] {$\rul^4_1$};
\node[rulenode,scale=0.5] (r8) [below right=\distb and \distb of r1] {$\rul^5_1$};

\node[rulenode,dotted,scale=0.5] (r2) [below right=-0.4 and 4cm of g1] {$\rul^1_4$};
\node[rulenode,dotted,scale=0.5] (r4) [below right=\distb and \distb of r2] {$\rul^1_3$};
\node[rulenode,dotted,scale=0.5] (r3) [below right=\distb and \distb of r4] {$\rul^1_2$};
\node[rulenode,scale=0.5] (r5) [below right=\distb and \distb of r3] {$\rul^1_5$};

\node[aggnode,scale=0.5] (rh1) [right= 0.3cm of r5] {A1};

\node[aggnode,dotted,scale=0.5] (rh2) [right= 0.3cm of r6] {A2};
\node[aggnode,dotted,scale=0.5] (rh3) [below right=\distb and \distb of rh2] {A3};
\node[aggnode,dotted,scale=0.5] (rh4) [below right=\distb and \distb of rh3] {A4};
\node[aggnode,scale=0.5] (rh5) [below right=\distb and \distb of rh4] {A5};

\draw[gray,->] (r1) -> (rh4);
\draw[gray,->] (rh4) -> (h4);

\draw[gray,->] (r8) -> (rh5);
\draw[gray,->] (rh3) -> (h3);

\draw[gray,->] (r2) -> (rh1);
\draw[gray,->] (r3) -> (rh1);
\draw[gray,->] (r4) -> (rh1);
\draw[gray,->] (r5) -> (rh1);
\draw[gray,->] (rh1) -> (h1);

\draw[gray,->] (r7) -> (rh3);
\draw[gray,->] (rh5) -> (h5);

\draw[gray,->] (r6) -> (rh2);
\draw[gray,->] (rh2) -> (h2);

\draw[black,dotted] (h1) -> (h2);
\draw[black,dotted] (h1) -> (h3);
\draw[black,dotted] (h1) -> (h5);
\draw[black,dotted] (h1) -> (h4);

\draw[violet,->] (g1) to[out=2,in=165] node[above] {$W_2$} (h1);

\draw[violet,->] (g2) to[out=-2,in=-165] node[below] {} (h2);
\draw[violet,->] (g4) to[out=-2,in=-165] node[below] {} (h4);
\draw[violet,->] (g3) to[out=-2,in=-165] node[below] {} (h3);
\draw[violet,->] (g5) to[out=-2,in=-165] node[below] {\textcolor{black}{$4\times$}$W_2$} (h5);

\draw[orange,->] (g1) -> node[below, near end] {\textcolor{black}{$4\times$}$W_1$} (r1);
\draw[orange,->] (g1) -> node[above, near end] {} (r6);
\draw[orange,->] (g1) -> node[above, near end] {} (r7);
\draw[orange,->] (g1) -> node[above, near end] {} (r8);

\draw[orange,->] (g2) -> node[above, near end] {\textcolor{black}{$4\times$}$W_1$} (r2);
\draw[orange,->] (g3) -> node[below, near end] {} (r4);
\draw[orange,->] (g4) -> node[below, near end] {} (r3);
\draw[orange,->] (g5) -> node[below, near end] {} (r5);

\node[recnode2] (hh1) [right = 5.5cm of h1] {$\rul_{h(c_1)}^{(n)}$};
\node[recnode2,dotted] (hh2) [below=1cm of hh1] {$\rul_{h(n_2)}^{(n)}$};
\node[recnode2,dotted] (hh3) [below right=\dist and \dista of hh2] {$\rul_{h(n_3)}^{(n)}$};
\node[recnode2,dotted] (hh4) [below right=\dist and \dista of hh3] {$\rul_{h(n_4)}^{(n)}$};
\node[recnode2] (hh5) [below right=\dist and \dista of hh4] {$\rul_{h(h_4)}^{(n)}$};

\draw[black,loosely dotted] (hh1) -> (hh2);
\draw[black,loosely dotted] (hh1) -> (hh3);
\draw[black,loosely dotted] (hh1) -> (hh4);
\draw[black,loosely dotted] (hh1) -> (hh5);

\draw[cyan,dashed,->] (h1) -> node[color=black, draw=white, fill=white] {\Large$\dots$} (hh1);
\draw[cyan,dashed,->] (h2) -> node[color=black, draw=white, fill=white] {\Large$\dots$} (hh2);
\draw[cyan,dashed,->] (h4) -> node[color=black, draw=white, fill=white] {\Large$\dots$} (hh4);
\draw[cyan,dashed,->] (h5) -> node[color=black, draw=white, fill=white] {\Large$\dots$} (hh5);
\draw[cyan,dashed,->] (h3) -> node[color=black, draw=white, fill=white] {\Large$\dots$} (hh3);

\draw[blue,dashed,->] (h1) -> node[color=black, draw=white, fill=white] {\Large$\dots$} (hh2);
\draw[blue,dashed,->] (h1) -> node[color=black, draw=white, fill=white] {\Large$\dots$} (hh3);
\draw[blue,dashed,->] (h1) -> node[color=black, draw=white, fill=white] {\Large$\dots$} (hh4);
\draw[blue,dashed,->] (h1) -> node[color=black, draw=white, fill=white] {\Large$\dots$} (hh5);

\draw[blue,dashed,->] (h2) -> node[color=black, draw=white, fill=white] {\Large$\dots$} (hh1);
\draw[blue,dashed,->] (h3) -> node[color=black, draw=white, fill=white] {\Large$\dots$} (hh1);
\draw[blue,dashed,->] (h4) -> node[color=black, draw=white, fill=white] {\Large$\dots$} (hh1);
\draw[blue,dashed,->] (h5) -> node[color=black, draw=white, fill=white] {\Large$\dots$} (hh1);

\node[aggnode] (agg) [right = 2cm of hh1] {$\agg^{\lab}$};
\draw[gray,->] (hh1) -> (agg);
\draw[gray,->] (hh2) -> (agg);
\draw[gray,->] (hh3) -> (agg);
\draw[gray,->] (hh4) -> (agg);
\draw[gray,->] (hh5) -> (agg);

\node[qnode] (q) [right = 1.5cm of agg] {$\rul^{\lab}$};
\draw[brown,->] (agg) -> node[below] {$W^{(n)}$} (q);

\end{tikzpicture}};
	  \node[scale=0.4] (met) at (9,-0.8){\begin{tikzpicture}
[transform shape,rotate=0, node distance=2.0cm and 2.0cm,
ar/.style={->,>=latex},
mynode/.style={
  draw, scale = 1.0,  minimum size=1cm, rounded corners,
  minimum height=1cm,
  align=center
  }
]

\node[atom, ball color=red!50!white] at (0,1.4) (o1) {\Large \textbf{C\textsubscript{{\tiny 1}}}};

\node[atom, circle, dotted, left color=blue!99!white] at (0.4,0) (h1) {\textbf{H\textsubscript{\tiny 2}}};
\node[atom, circle, dotted, left color=blue!99!white] at (0.15,0) (h2) {\textbf{H\textsubscript{\tiny 3}}};
\node[atom, circle, dotted, left color=blue!99!white] at (-0.15,0) (h2) {\textbf{H\textsubscript{\tiny 1}}};
\node[atom, circle, ball color=blue!50!white] at (-0.4,0) (h2) {\textbf{H\textsubscript{\tiny 4}}};

\node[mynode, dashed] at (4,1.3) (q) {\Large\textbf{$\lab$}};

\draw[black, -{Latex[length=5mm, width=4mm]}, dashed, shorten <=0.9cm] (o1) -> (q);

\end{tikzpicture}};
	\end{tikzpicture}
}
\caption{A compressed version of the GNN from Fig.~\ref{fig:GNNs}, with the compressed parts~dotted.}
\label{fig:groundCompressed}
\end{figure*}

\subsection{An Exact Compression Algorithm}\label{sec:exact-algorithm}

The exact algorithm for compressing computation graphs reuses the evaluation with random parameter initializations while recording the respective values for all the nodes. However, the next steps are different. First, instead of traversing the computation graph from the output nodes towards the leaves, it traverses the graph bottom-up, starting from the leaves. Second, rather than merging the nodes with the same recorded value lists right away, the exact algorithm merely considers these as candidates for merging. For that it keeps a data structure (based on a hash table) that indexes the nodes of the computation graph by the lists of the respective values recorded for the random parameter initializations. When, while traversing the graph, it processes a node $N$, it checks if there is any node $N’$ that had the same values over all the random parameter initializations and has already been processed. If so it checks if $N$ and $N’$ are {\em structurally equivalent} (which we explain in turn) and if they are it replaces $N$ by $N’$. 
To test the structural equivalence of two nodes, the algorithm checks the following conditions:

\begin{enumerate}
    \item The activation functions of $N$ and $N'$ are the same.
    \item The lists of children of both $N$ and $N'$ are the same (not just structurally equivalent but identical, i.e. $\textit{Children}(N) = \textit{Children}(N')$), and if $C$ is the $i$-th child of $N$ and $C'$ is the $i$-th child of $N'$, with $(C,N,L_1)$ and $(C',N,L_2)$ being the respective edges connecting them to $N$, then the labels $L_1$ and $L_2$ must be equal, too. 
\end{enumerate}


One can show why the above procedure works by induction. We sketch the main idea here. The base case is trivial. To show the inductive step we can reason as follows. When we are processing the node $N$, by the assumption, the node $N'$ has already been processed. Thus, we know that the children of both $N$ and $N'$ must have already been processed as well. By the induction hypothesis, if any of the children were structurally equivalent, they must have been merged by the algorithm, and so it is enough to check identity of the child nodes. This reasoning then allows one to easily finish a proof of correctness of this algorithm.


There is one additional optimization that we can do for symmetric activation functions. Here by ``symmetric'' we mean symmetric with respect to permutation of the arguments. An example of such a symmetric activation function is any function of the form $f(x_1,\dots,x_k) = h\left(\sum_{i=1}^k\ x_k \right)$; such functions are often used in neural networks. In this case we replace the condition 2 above by:

\begin{enumerate}
    \item[2'.] There is a permutation $\pi$ such that $\pi(\textit{Children}(N)) = \textit{Children}(N')$), and if $C$ is the $i$-th child of $N$ and $C'$ is the $\pi(i)$-th child of $N'$, with $(C,N,L_1)$ and $(C',N,L_2)$ being the respective edges connecting them to $N$, then the labels $L_1$ and $L_2$ must be equal. 
\end{enumerate}

It is not difficult to implement the above check efficiently (we omit the details for brevity). Note also that the overall asymptotic complexity of compressing a graph $\mathcal{G}$ with either of the algorithms is simply linear in the size of the graph. Specifically, it is the same as the $n$ evaluations of $\mathcal{G}$.


Finally, to illustrate the effect of the lossless compression, we show the GNN model (Sec.~\ref{sec:gnn}), unfolded over a sample molecule of methane from~Fig.\ref{fig:GNNs}, compressed in Fig.~\ref{fig:groundCompressed}.

\section{Experiments}
\label{sec:experiments}


To test the proposed compression in practice, we selected some common structured convolutional models, and evaluated them on a number of real datasets from the domains of (i) molecule classification and (ii) knowledge-base completion. The questions to be~answered~by~the~experiments~are:

\begin{enumerate}
    \item How numerically efficient is the non-exact algorithm in achieving lossless compression?
    \item What improvements does the compression provide in terms of graph size and speedup?
    \item Is learning accuracy truly unaffected by the, presumably lossless, compression in practice?
\end{enumerate}

\paragraph{Models}
We chose mostly GNN-based models as their dynamic computation graphs encompass all the elements of structured convolutional models (convolution, pooling, and recursive layer stacking). Particularly, we choose well-known instances of GCNs and graph-SAGE (Sec.~\ref{sec:gnn}), each with 2 layers. Additionally, we include Graph Isomorphism Networks (GIN)~\citep{xu2018powerful}, which follow the same computation scheme with 5 layers, but their particular operations ($C_{W_1}=identity$,~$A=sum$,~$C_{W_2}=MLP$) are theoretically substantiated in the expressiveness of the Weisfeiler-Lehman test~\citep{weisfeiler2006construction}. This is interesting in that it should effectively distinguish non-isomorphic substructures in the data by generating consistently distinct computations, and should thus be somewhat more resistant to our proposed compression. Finally, we include a relational template (``graphlets'') introduced in~\citep{sourek2018lifted}, which generalizes GNNs to aggregate small 3-graphlets instead of just neighbors.

\paragraph{Datasets}
For structure property prediction, we used 78 organic molecule classification datasets reported in previous works~\citep{ncigi,PTC,mutagenesis}. Nevertheless, we show only the (alphabetically) first 3 for clarity, as the target metrics were extremely similar over the whole set.
We note we also extended GCNs with edge embeddings to account for the various bond types, further \textit{decreasing} the symmetries.
For knowledge base completion (KBC), we selected commonly known datasets of Kinships, Nations, and UMLS~\citep{StatisticalPredicateInventionPaper} composed of different object-predicate-subject triplets. We utilized GCNs to learn embeddings of all the items and relations, similarly to R-GCNs~\citep{schlichtkrull2018modeling}, and for prediction of each triplet, we fed the three embeddings into an MLP, such as in~\citep{dong2014knowledge}, denoted as ``KBE''.

The size of the individual input graphs is generally smallest in the molecular data with app. 25 atoms and 50 bonds per a single molecule, where there are app. 3000 molecules in each of the datasets on average. The input graphs are then naturally largest for the knowledge bases with app. 20,000 triples over hundreds of objects and relations. The sizes of the corresponding computation graphs themselves are then in the orders of $10^2$--$10^5$ nodes, respectively.

\begin{figure*}[]
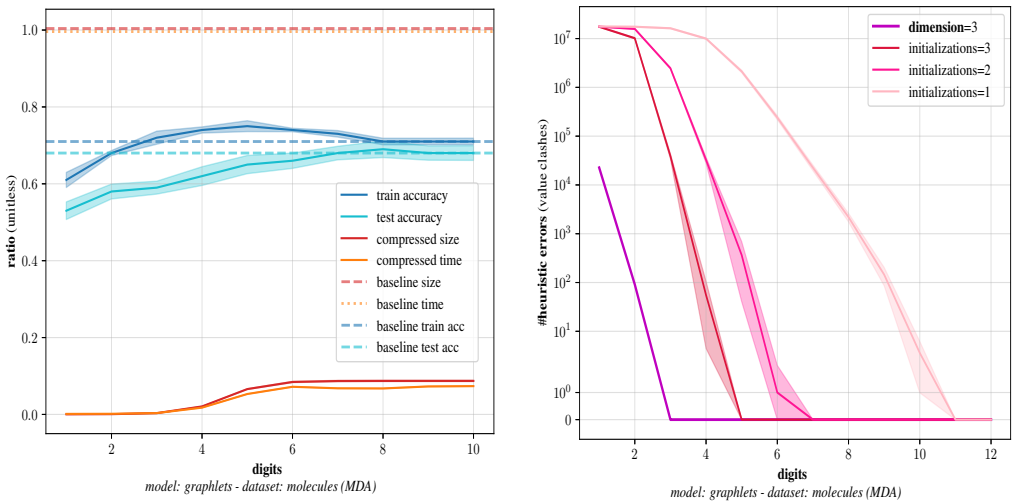

\centering
\resizebox{1.03\linewidth}{7cm}{
\input{img/graphs/iso_scalar_all_new.pgf}
\raisebox{4mm}[0pt][0pt]{
\input{img/graphs/isocheck_combined_new.pgf}
}
}
\caption{
Compression of a \textit{scalar}-parameterized graphlets model on a molecular dataset. We display progression of the selected metrics w.r.t. increasing number of significant digits (inits=$1$) used in the value comparisons (left), and number of non-equivalent subgraph value clashes detected by the exact algorithm w.r.t. the digits, weight re-initializations, and increased weight dimension (right).
}
\label{fig:iso_scalar_digits}
\end{figure*}

\paragraph{Experimental Protocol}
We approached all the learning scenarios under simple unified setting with standard hyperparameters, none of which was set to help the compression (sometimes on~the~contrary). We used the (re-implemented) LRNN framework to encode all the models, and also compared with popular GNN frameworks of PyTorch Geometric (PyG)~\citep{fey2019fast} and Deep Graph Library (DGL)~\citep{wang2019deep}. If not dictated by the particular model, we set the activation functions simply as $C_W=\frac{1}{1+e^{-W\cdot x}}$ and $A=avg$. We then trained against $MSE$ using $1000$ steps of ADAM, and evaluated with a $5$-fold crossvalidation.


\begin{table}[]
\centering
\caption{Training times \textit{per epocha} across different models and frameworks over 3000 molecules. Additionally, the startup graphs creation time of~LRNNs (including the compression)~is~reported.}
\begin{tabular}{c|c|c|c|c|c}
Model & Lifting (s) & LRNNs (s)  & PyG (s)        & DGL (s)        & LRNN startup (s) \\
\hline
\hline
GCN         & \textbf{0.25 $\pm$ 0.01} & 0.75$\pm$ 0.01 & 3.24 $\pm$ 0.02  & 23.25 $\pm$ 1.94 & 35.2 $\pm$ 1.3        \\
\hline
g-SAGE      & \textbf{0.34 $\pm$ 0.01} & 0.89$\pm$ 0.01 & 3.83 $\pm$ 0.04  & 24.23 $\pm$ 3.80 & 35.4 $\pm$ 1.8        \\
\hline
GIN         & \textbf{1.41 $\pm$ 0.10} & 2.84$\pm$ 0.09 & 11.19 $\pm$ 0.06 & 52.04 $\pm$ 0.41 & 75.3 $\pm$ 3.2         
\end{tabular}
\label{tab:times}
\end{table}

\subsection{Results}


Firstly, we tested numerical efficiency of the non-exact algorithm itself (Sec.~\ref{sec:lifting}), for which we used scalar weight representation in the models to detect symmetries on the level of individual ``neurons'' (rather than ``layers''). We used the (most expressive) graphlets model, where we checked the functional symmetries to overlap with the structural symmetries. The results in Fig.~\ref{fig:iso_scalar_digits} then show that the non-exact algorithm is already able to perfectly distinguish all structural symmetries with but a single weight initialization within less than 12 significant digits. While more initializations indeed improved the efficiency rapidly, in the end they proved unnecessary (but could be used in cases where the available precision would be insufficient).
Moreover this test was performed with the actual low-range logistic activations. The displayed ($10$x) training time improvement (Fig.\ref{fig:iso_scalar_digits} - left) in the scalar models was then directly reflecting the network size reduction, and could be pushed further by decreasing the numeric precision at the expected cost of degrading the learning performance.

\begin{figure*}[t]
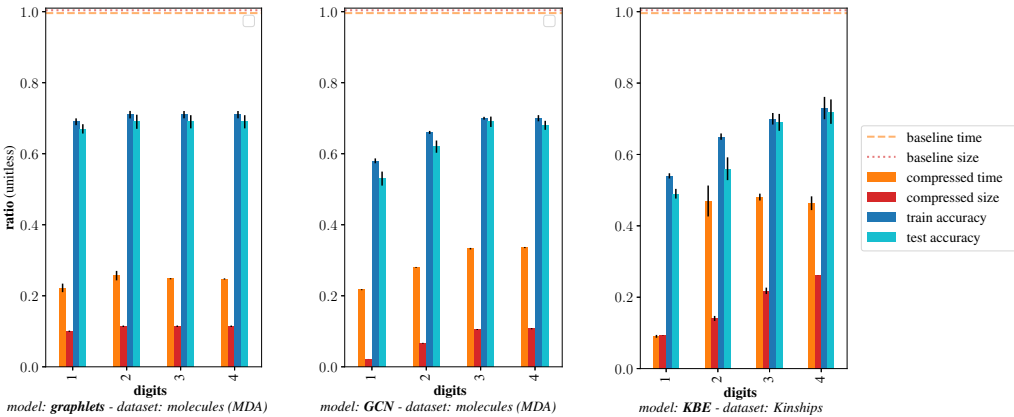

\centering
\resizebox{1.0\linewidth}{!}{
\input{img/graphs/iso_vector_digits_lrnn.pgf}
\input{img/graphs/iso_vector_digits_gnn.pgf}
\input{img/graphs/iso_vector_digits_kinships.pgf}
}
\caption{Compression of 3 \textit{tensor}-parameterized models of graphlets (left), GCNs (middle) and KBEs (right) over the molecular (left, middle) and Kinships (right) datasets, with progression of selected metrics against the increasing number of significant digits used for equivalence checking. 
}
\label{fig:iso_vector_digits}
\end{figure*}

Secondly, we performed similar experiments with standard tensor parameterization, where the equivalences were effectively detected on the level of whole neural ``layers'', since the vector output values (of dim=3)
were compared for equality instead. This further improved the precision of the non-exact algorithm (Fig.~\ref{fig:iso_scalar_digits} - right), where merely the first 4 digits were sufficient to achieve lossless compression in all the models and datasets (Figure~\ref{fig:iso_vector_digits}). However, the training (inference) time was no longer directly reflecting the network size reduction, which we account to optimizations used in the vectorized computations. Nevertheless the speedup (app. $3$x) was still substantial.

\begin{figure}
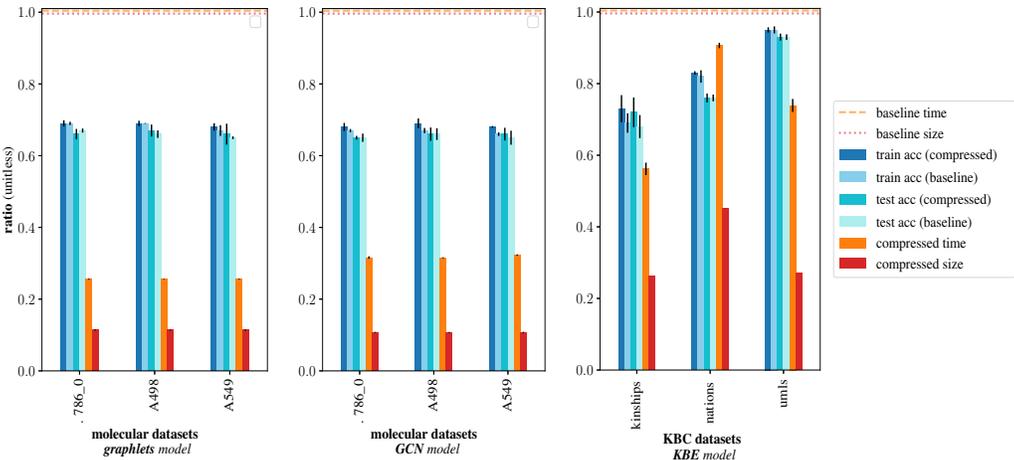

\centering
\resizebox{1.0\linewidth}{6.4cm}{
\input{img/graphs/iso_vector_lossless_lrnn.pgf}
\input{img/graphs/iso_vector_lossless_gnn.pgf}
\raisebox{-2mm}[0pt][0pt]{
\input{img/graphs/iso_vector_lossless_kbc.pgf}
}
}
\caption{Comparison of 3 different baseline models of graphlets (left), GCNs (middle), and KBEs (right)  with their compressed versions over molecule classification (left, middle) and KBC (right).
}
\label{fig:iso_onoff}
\end{figure}

We further compared with established GNN frameworks of PyG~\citep{fey2019fast} and DGL~\citep{wang2019deep}. We made sure to align the exact computations of GCN, graph-SAGE, and GIN, while all the frameworks performed equally w.r.t. the accuracies. For a more fair comparison, we further increased all (tensor) dimensions to a more common dim=10. The compression effects, as well as performance edge of the implemented LRNN framework itself, are displayed in Tab.~\ref{tab:times} for a sample molecular dataset (MDA). Note that the compression was truly least effective for the aforementioned GIN model, nevertheless still provided app. $2$x speedup.

Finally, the results in Fig.~\ref{fig:iso_onoff} confirm that the proposed lossless compression via lifting, with either the exact algorithm or the non-exact algorithm with a high-enough numeric precision used, indeed does not degrade the learning performance in terms of training and testing accuracy (both were close within margin of variance over the crossvalidation folds).


Note that the used templated models are quite simple and do not generate any symmetries on their own (which they would, e.g., with recursion), but rather merely reflect the symmetries in the data themselves. Consequently, the speedup was overall lowest for the knowledge graph of Nations, represented via 2 distinct relation types, and higher for the Kinships dataset, representing a more densely interconnected social network. The improvement was then generally biggest for the highly symmetric molecular graphs where, interestingly, the compression often reduced the neural computation graphs to a size even smaller than that of the actual input molecules. Note we only compressed symmetries within individual computation graphs (samples), and the results thus cannot be biased by the potential existence of isomorphic samples~\citep{ivanov2019understanding}, however, potentially much higher compression rates could be also achieved with (dynamic) batching.

\section{Conclusion}

We introduced a simple, efficient, lossless compression technique for structured convolutional models inspired by lifted inference. The technique is very light-weight and can be easily adopted by any neural learner, but is most effective for structured convolutional models utilizing weight sharing schemes to target relational data, such as in various graph neural networks. We have demonstrated with existing models and datasets that a significant inference and training time reduction can be achieved without affecting the learning results, and possibly extended beyond for additional speedup.

\subsection*{Acknowledgments}

This work was supported by the Czech Science Foundation project GJ20-19104Y. GS and FZ are also supported by the Czech Science Foundation project 20-29260S.




\bibliography{lifting}

\begin{thebibliography}{33}
\providecommand{\natexlab}[1]{#1}
\providecommand{\url}[1]{\texttt{#1}}
\expandafter\ifx\csname urlstyle\endcsname\relax
  \providecommand{\doi}[1]{doi: #1}\else
  \providecommand{\doi}{doi: \begingroup \urlstyle{rm}\Url}\fi

\bibitem[Cheng et~al.(2017)Cheng, Wang, Zhou, and Zhang]{cheng2017survey}
Yu~Cheng, Duo Wang, Pan Zhou, and Tao Zhang.
\newblock A survey of model compression and acceleration for deep neural
  networks.
\newblock \emph{arXiv preprint arXiv:1710.09282}, 2017.

\bibitem[DeMillo \& Lipton(1977)DeMillo and Lipton]{demillo1977probabilistic}
Richard~A DeMillo and Richard~J Lipton.
\newblock A probabilistic remark on algebraic program testing.
\newblock Technical report, Georgia Inst. of Technology, Atlanta School of
  Information and Computer science, 1977.

\bibitem[Dong et~al.(2014)Dong, Gabrilovich, Heitz, Horn, Lao, Murphy,
  Strohmann, Sun, and Zhang]{dong2014knowledge}
Xin Dong, Evgeniy Gabrilovich, Geremy Heitz, Wilko Horn, Ni~Lao, Kevin Murphy,
  Thomas Strohmann, Shaohua Sun, and Wei Zhang.
\newblock Knowledge vault: A web-scale approach to probabilistic knowledge
  fusion.
\newblock In \emph{Proceedings of the 20th ACM SIGKDD international conference
  on Knowledge discovery and data mining}, pp.\  601--610, 2014.

\bibitem[Fey \& Lenssen(2019)Fey and Lenssen]{fey2019fast}
Matthias Fey and Jan~Eric Lenssen.
\newblock Fast graph representation learning with pytorch geometric.
\newblock \emph{arXiv preprint arXiv:1903.02428}, 2019.

\bibitem[Gilmer et~al.(2017)Gilmer, Schoenholz, Riley, Vinyals, and
  Dahl]{gilmer2017neural}
Justin Gilmer, Samuel~S Schoenholz, Patrick~F Riley, Oriol Vinyals, and
  George~E Dahl.
\newblock Neural message passing for quantum chemistry.
\newblock In \emph{Proceedings of the 34th International Conference on Machine
  Learning-Volume 70}, pp.\  1263--1272. JMLR. org, 2017.

\bibitem[Hamilton et~al.(2017)Hamilton, Ying, and
  Leskovec]{hamilton2017inductive}
Will Hamilton, Zhitao Ying, and Jure Leskovec.
\newblock Inductive representation learning on large graphs.
\newblock In \emph{Advances in neural information processing systems}, pp.\
  1024--1034, 2017.

\bibitem[Helma et~al.(2001)Helma, King, Kramer, and Srinivasan]{PTC}
Christoph Helma, Ross~D. King, Stefan Kramer, and Ashwin Srinivasan.
\newblock The predictive toxicology challenge 2000--2001.
\newblock \emph{Bioinformatics}, 17\penalty0 (1):\penalty0 107--108, 2001.

\bibitem[Ivanov et~al.(2019)Ivanov, Sviridov, and
  Burnaev]{ivanov2019understanding}
Sergei Ivanov, Sergei Sviridov, and Evgeny Burnaev.
\newblock Understanding isomorphism bias in graph data sets.
\newblock \emph{arXiv preprint arXiv:1910.12091}, 2019.

\bibitem[Kersting(2012)]{kersting2012lifted}
Kristian Kersting.
\newblock Lifted probabilistic inference.
\newblock In \emph{ECAI}, pp.\  33--38, 2012.

\bibitem[Kersting \& {De Raedt}(2001)Kersting and {De Raedt}]{BLP}
Kristian Kersting and Luc {De Raedt}.
\newblock Towards combining inductive logic programming with bayesian networks.
\newblock In \emph{Inductive Logic Programming, 11th International Conference,
  {ILP} 2001, Strasbourg, France, September 9-11, 2001, Proceedings}, pp.\
  118--131, 2001.

\bibitem[Kimmig et~al.(2015)Kimmig, Mihalkova, and Getoor]{LIFTED}
A~Kimmig, L~Mihalkova, and L~Getoor.
\newblock Lifted graphical models: a survey.
\newblock \emph{Machine Learning}, 99\penalty0 (1):\penalty0 1--45, 2015.

\bibitem[Kipf \& Welling(2016)Kipf and Welling]{kipf2016semi}
Thomas~N Kipf and Max Welling.
\newblock Semi-supervised classification with graph convolutional networks.
\newblock \emph{arXiv preprint arXiv:1609.02907}, 2016.

\bibitem[Kok \& Domingos(2007)Kok and
  Domingos]{StatisticalPredicateInventionPaper}
Stanley Kok and Pedro Domingos.
\newblock Statistical predicate invention.
\newblock In \emph{Proceedings of the 24th International Conference on Machine
  Learning}, pp.\  433--440, 2007.

\bibitem[Koller et~al.(2007)Koller, Friedman, D{\v{z}}eroski, Sutton, McCallum,
  Pfeffer, Abbeel, Wong, Heckerman, Meek, et~al.]{koller2007introduction}
Daphne Koller, Nir Friedman, Sa{\v{s}}o D{\v{z}}eroski, Charles Sutton, Andrew
  McCallum, Avi Pfeffer, Pieter Abbeel, Ming-Fai Wong, David Heckerman, Chris
  Meek, et~al.
\newblock \emph{Introduction to statistical relational learning}.
\newblock MIT press, 2007.

\bibitem[Krizhevsky et~al.(2012)Krizhevsky, Sutskever, and Hinton]{CNNS}
Alex Krizhevsky, Ilya Sutskever, and Geoffrey~E Hinton.
\newblock Imagenet classification with deep convolutional neural networks.
\newblock In \emph{Advances in neural information processing systems}, pp.\
  1097--1105, 2012.

\bibitem[Lodhi \& Muggleton(2005)Lodhi and Muggleton]{mutagenesis}
Huma Lodhi and Stephen Muggleton.
\newblock Is mutagenesis still challenging.
\newblock \emph{ILP-Late-Breaking Papers}, 35, 2005.

\bibitem[Manhaeve et~al.(2018)Manhaeve, Dumancic, Kimmig, Demeester, and
  De~Raedt]{manhaeve2018deepproblog}
Robin Manhaeve, Sebastijan Dumancic, Angelika Kimmig, Thomas Demeester, and Luc
  De~Raedt.
\newblock Deepproblog: Neural probabilistic logic programming.
\newblock In \emph{Advances in Neural Information Processing Systems}, pp.\
  3749--3759, 2018.

\bibitem[Marra \& Ku{\v{z}}elka(2019)Marra and Ku{\v{z}}elka]{marra2019neural}
Giuseppe Marra and Ond{\v{r}}ej Ku{\v{z}}elka.
\newblock Neural markov logic networks.
\newblock \emph{arXiv preprint arXiv:1905.13462}, 2019.

\bibitem[Ralaivola et~al.(2005)Ralaivola, Swamidass, Saigo, and Baldi]{ncigi}
Liva Ralaivola, Sanjay~J. Swamidass, Hiroto Saigo, and Pierre Baldi.
\newblock Graph kernels for chemical informatics.
\newblock \emph{Neural Netw.}, 18\penalty0 (8):\penalty0 1093--1110, 2005.

\bibitem[Richardson \& Domingos(2006)Richardson and Domingos]{MLN}
Matthew Richardson and Pedro Domingos.
\newblock Markov logic networks.
\newblock \emph{Machine learning}, 2006.

\bibitem[Rockt{\"a}schel \& Riedel(2017)Rockt{\"a}schel and
  Riedel]{rocktaschel2017end}
Tim Rockt{\"a}schel and Sebastian Riedel.
\newblock End-to-end differentiable proving.
\newblock In \emph{Advances in Neural Information Processing Systems}, 2017.

\bibitem[Schlichtkrull et~al.(2018)Schlichtkrull, Kipf, Bloem, Van Den~Berg,
  Titov, and Welling]{schlichtkrull2018modeling}
Michael Schlichtkrull, Thomas~N Kipf, Peter Bloem, Rianne Van Den~Berg, Ivan
  Titov, and Max Welling.
\newblock Modeling relational data with graph convolutional networks.
\newblock In \emph{European Semantic Web Conference}, pp.\  593--607. Springer,
  2018.

\bibitem[Sen et~al.(2012)Sen, Deshpande, and Getoor]{sen2012bisimulation}
Prithviraj Sen, Amol Deshpande, and Lise Getoor.
\newblock Bisimulation-based approximate lifted inference.
\newblock \emph{arXiv preprint arXiv:1205.2616}, 2012.

\bibitem[Serra et~al.(2020)Serra, Kumar, and Ramalingam]{serra2020lossless}
Thiago Serra, Abhinav Kumar, and Srikumar Ramalingam.
\newblock Lossless compression of deep neural networks.
\newblock \emph{arXiv preprint arXiv:2001.00218}, 2020.

\bibitem[Socher et~al.(2013)Socher, Perelygin, Wu, Chuang, Manning, Ng, Potts,
  et~al.]{socher2013recursive}
Richard Socher, Alex Perelygin, Jean~Y Wu, Jason Chuang, Christopher~D Manning,
  Andrew~Y Ng, Christopher Potts, et~al.
\newblock Recursive deep models for semantic compositionality over a sentiment
  treebank.
\newblock In \emph{Proceedings of the conference on empirical methods in
  natural language processing (EMNLP)}, volume 1631, pp.\  1642. Citeseer,
  2013.

\bibitem[Sourek et~al.(2018)Sourek, Aschenbrenner, Zelezny, Schockaert, and
  Kuzelka]{sourek2018lifted}
Gustav Sourek, Vojtech Aschenbrenner, Filip Zelezny, Steven Schockaert, and
  Ondrej Kuzelka.
\newblock Lifted relational neural networks: Efficient learning of latent
  relational structures.
\newblock \emph{Journal of Artificial Intelligence Research}, 62:\penalty0
  69--100, 2018.

\bibitem[Suciu et~al.(2011)Suciu, Olteanu, R{\'e}, and
  Koch]{suciu2011probabilistic}
Dan Suciu, Dan Olteanu, Christopher R{\'e}, and Christoph Koch.
\newblock Probabilistic databases.
\newblock \emph{Synthesis lectures on data management}, 3\penalty0
  (2):\penalty0 1--180, 2011.

\bibitem[Wang et~al.(2019)Wang, Yu, Zheng, Gan, Gai, Ye, Li, Zhou, Huang, Ma,
  et~al.]{wang2019deep}
Minjie Wang, Lingfan Yu, Da~Zheng, Quan Gan, Yu~Gai, Zihao Ye, Mufei Li,
  Jinjing Zhou, Qi~Huang, Chao Ma, et~al.
\newblock Deep graph library: Towards efficient and scalable deep learning on
  graphs.
\newblock \emph{arXiv preprint arXiv:1909.01315}, 2019.

\bibitem[Wang et~al.(2017)Wang, Mao, Wang, and Guo]{wang2017knowledge}
Quan Wang, Zhendong Mao, Bin Wang, and Li~Guo.
\newblock Knowledge graph embedding: A survey of approaches and applications.
\newblock \emph{IEEE Transactions on Knowledge and Data Engineering},
  29\penalty0 (12):\penalty0 2724--2743, 2017.

\bibitem[Weisfeiler(2006)]{weisfeiler2006construction}
Boris Weisfeiler.
\newblock \emph{On construction and identification of graphs}, volume 558.
\newblock Springer, 2006.

\bibitem[Wu et~al.(2019)Wu, Pan, Chen, Long, Zhang, and
  Yu]{wu2019comprehensive}
Zonghan Wu, Shirui Pan, Fengwen Chen, Guodong Long, Chengqi Zhang, and Philip~S
  Yu.
\newblock A comprehensive survey on graph neural networks.
\newblock \emph{arXiv preprint arXiv:1901.00596}, 2019.

\bibitem[Xu et~al.(2018{\natexlab{a}})Xu, Hu, Leskovec, and
  Jegelka]{xu2018powerful}
Keyulu Xu, Weihua Hu, Jure Leskovec, and Stefanie Jegelka.
\newblock How powerful are graph neural networks?
\newblock \emph{arXiv preprint arXiv:1810.00826}, 2018{\natexlab{a}}.

\bibitem[Xu et~al.(2018{\natexlab{b}})Xu, Li, Tian, Sonobe, Kawarabayashi, and
  Jegelka]{xu2018representation}
Keyulu Xu, Chengtao Li, Yonglong Tian, Tomohiro Sonobe, Ken-ichi Kawarabayashi,
  and Stefanie Jegelka.
\newblock Representation learning on graphs with jumping knowledge networks.
\newblock \emph{arXiv preprint arXiv:1806.03536}, 2018{\natexlab{b}}.

\end{thebibliography}
\bibliographystyle{iclr2021_conference}
\end{document}